\documentclass[a4paper, 10pt, conference]{ieeeconf}      
\usepackage{FG2020}
\usepackage{times}
\usepackage{epsfig}
\usepackage{graphicx}
\usepackage{amsmath}
\usepackage{amssymb}
\usepackage{multirow}
\usepackage{booktabs}
\usepackage{algorithm}
\usepackage[noend]{algpseudocode}
\usepackage{url}
\usepackage{subfigure}
\usepackage{dblfloatfix}
\usepackage{calc}
\usepackage{amstext}
\usepackage{multicol}
\usepackage{pslatex}

\usepackage[font={small}]{caption}
\newcommand{\RN}[1]{%
  \textup{\uppercase\expandafter{\romannumeral#1}}%
}

\FGfinalcopy 

\IEEEoverridecommandlockouts                              
\def\FGPaperID{****} 

\title{\LARGE \bf
DriverMHG: A Multi-Modal Dataset for Dynamic Recognition of Driver Micro Hand Gestures and a Real-Time Recognition Framework
}

\author{\parbox{16cm}{\centering
    {\large Okan K\"op\"ukl\"u$^1$, Thomas Ledwon$^2$, Yao Rong$^3$, Neslihan Kose$^4$, Gerhard Rigoll$^1$}\\
    {\normalsize
    \vspace{0.2cm}
    $^1$ Institute for Human-Machine Communication, TU Munich, Germany\\
    $^2$ Institute of Informatics, LMU Munich, Germany\\
    $^3$ Chair of Human-Computer Interaction, University of Tuebingen, Germany\\
    $^4$ Dependability Research Lab, Intel Labs Europe, Intel Deutschland GmbH, Germany}}
}

\begin{document}

\ifFGfinal
\pagestyle{plain}
\else
\author{Anonymous FG2020 submission\\ Paper ID \FGPaperID \\}
\pagestyle{plain}
\fi
\maketitle
\thispagestyle{plain}

\begin{abstract}

The use of hand gestures provides a natural alternative to cumbersome interface devices for Human-Computer Interaction (HCI) systems. However, real-time recognition of dynamic micro hand gestures from video streams is challenging for in-vehicle scenarios since (i) the gestures should be performed naturally without distracting the driver, (ii) micro hand gestures occur within very short time intervals at spatially constrained areas, (iii) the performed gesture should be recognized only once, and (iv) the entire architecture should be designed lightweight as it will be deployed to an embedded system. In this work, we propose an HCI system for dynamic recognition of driver micro hand gestures, which can have a crucial impact in automotive sector especially for safety related issues. For this purpose, we initially collected a dataset named Driver Micro Hand Gestures (DriverMHG), which consists of RGB, depth and infrared modalities. The challenges for dynamic recognition of micro hand gestures have been addressed by proposing a lightweight convolutional neural network (CNN) based architecture which operates online efficiently with a sliding window approach. For the CNN model, several 3-dimensional resource efficient networks are applied and their performances are analyzed. Online recognition of gestures has been performed with 3D-MobileNetV2, which provided the best offline accuracy among the applied networks with similar computational complexities. The final architecture is deployed on a driver simulator operating in real-time. We make DriverMHG dataset and our source code publicly available \footnote{https://www.mmk.ei.tum.de/DriverMHG/}.
\end{abstract}
\section{Introduction}

Computers have become an indispensable part of human life. Thus, facilitating natural human-computer interaction (HCI) contains utmost importance to bridge human-computer barrier. Although there is a growing interest in the development of new approaches and technologies for HCI, gestures have long been deemed to be a more natural, creative and intuitive interaction technique for communicating with our computers. 

\vspace{0.15cm}

In this work, we create an HCI system which is based on dynamic recognition of driver's micro hand gestures. In automotive sector, this kind of system can have a crucial impact especially for safety related issues. While driving, performing a hand gesture, which represents one action, by keeping the hands on the wheel is much safer than pressing a button to activate that action, which causes eyes off the road for few seconds. For this objective, the following challenges should be taken into account:

\begin{enumerate}
    \item A suitable dataset must be collected. The gestures in the dataset should be natural and should not distract the driver while performing. 
    \item The created architecture should distinguish the other hand movements when the driver is not performing any gesture.
    \item The architecture should be able to capture micro gestures, which are occurring within very short time intervals at spatially constrained areas, with an acceptable accuracy.
    \item The entire architecture should be designed considering the memory and power budget.
\end{enumerate}

Considering the aforementioned challenges, we initially collected a multi-modal micro hand gesture dataset using a driver simulator with 25 participants, who performed pre-defined micro gestures. This dataset is collected with one sensor providing synchronized RGB, infrared and depth modalities. To the best of our knowledge, this is the first multi-modal dataset that consists of micro hand gestures performed on a steering wheel.

Today, several video cameras provide more than one modality, and widely used ones are cameras providing RGB, infrared and depth modalities. Each modality has advantages, e.g. infrared is invariant to illumination and depth modality provides depth information. In this work, in addition to the mono-modal analysis with RGB, infrared and depth modalities, we have also analyzed the impact of fusion on the recognition performance of micro hand gestures.  

In real-world applications, resource efficiency, fast reaction time and single time activation are as crucial as reaching an acceptable accuracy for the created HCI system. In this work, we have applied several resource efficient \mbox{3-dimensional} (3D) CNN architectures proposed in \cite{kopuklu2019resource} as the CNN model of our dynamic micro hand gesture recognition approach. Among these architectures, 3D-MobileNetV2 has provided the best offline accuracy compared to the architectures with similar computational complexities. Therefore, in this paper online recognition analysis has been performed with 3D-MobileNetV2 as the CNN model. For online recognition, we have also proposed a novel single time activation approach, which does not require a separate \textit{gesture detector} block as in \cite{kopuklu2019real}.

In the proposed architecture, the video stream is split into two branches each containing only one hand. Then CNN models are trained separately on each hand and deployed to the online recognition architecture. For evaluating online recognition, we have used a recently proposed Levenshtein accuracy \cite{kopuklu2019real}. The experiments show that 3D-MobileNetV2 can operate online efficiently with a sliding window approach for dynamic micro hand gesture recognition. However, achieved 74.00\% online (Levenshtein) accuracy compared to 91.56\% offline accuracy shows that online recognition of micro gestures is challenging and open to improvements.

The rest of the paper is organized as follows. In Section \ref{ch:Related Work}, the related work in the area of action and gesture recognition is presented. Section \ref{sec:dataset} introduces the specifications of the driver micro hand gesture dataset. Section \ref{sec:approach} presents the proposed approach for dynamic recognition of micro hand gestures. Section \ref{sec:exp} shows the experiments and results. Finally, Section \ref{sec:conclusion} concludes the paper.
\section{Related Work}
\label{ch:Related Work}
Vision based recognition such as hand gesture recognition and action recognition have similar requirements. So far, different machine learning methods have been used for recognition tasks. One approach is to apply Hidden Markov Models (HMMs) \cite{HMM}, which are used in dynamic situations containing temporal information. Another one used for recognition tasks is k-Nearest Neighbors (k-NN) \cite{knn} classifier, which is simpler to implement. Support Vector Machine (SVM) \cite{SVM} is another commonly used and well-known classifier for gesture and action recognition, which maps the non-linear input data to a higher dimension, where the data is then linearly separated \cite{Suarez2012, Rautaray2015}. The more recent methods for recognition tasks are improved dense trajectories \cite{wang2016robust} and super normal vector \cite{yang2014super}. 

In 2012, AlexNet \cite{NIPS2012} achieved state-of-the-art classification accuracy at ImageNet challenge \cite{deng2009imagenet} against all the traditional machine learning approaches. It was a breakthrough for vision based recognition tasks and is the point in history where the interest in deep learning increased rapidly \cite{DLApproaches2018}. Today, vision based recognition approaches are dominated by the use of CNNs. After their success on image recognition, they have been explored also for video analysis tasks. A lot of work has proven that deep CNNs are capable to handle action recognition \cite{karpathy2014large, tran2015learning, zhang2016real} and hand gesture recognition \cite{Molchanov_2015_CVPR_Workshops, neverova2014multi, kopuklu2018motion} from video streams as well. 

First studies in action and gesture recognition field focused mostly on improving the classification accuracy from video streams with the application of deeper CNN architectures. Another approach commonly applied to improve accuracy with CNNs is the fusion of different modalities. CNNs are very suitable for fusing the information from several data modalities. So far, different fusion strategies have been explored for this purpose. In \cite{simonyan2014two, wang2016temporal}, two streams conveying temporal and spatial information from two separate CNNs are fused together for action classification. In \cite{kopuklu2018motion}, a data level fusion strategy, Motion Fused Frames (MFFs), is proposed. The advantage of this fusion strategy is that a single network training is sufficient, which reduces the number of parameters multiple times and any CNN architecture can be adapted with a very little modification. In all these works \cite{simonyan2014two, wang2016temporal, kopuklu2018motion}, the temporal information is extracted using optical flow, which is not very efficient for real-time applications due to its computational cost.

Today, there are several sensors that provide different modalities such as infrared, RGB and depth simultaneously. Since the input of CNNs is not only limited to RGB images, the architectures can be adapted for these modalities as well, and then they can be directly applied as input to CNNs without any additional computation. In \cite{yang2014super}, depth images are used for extending the surface normal for human activity recognition. In \cite{molchanov2015multi}, a specific recognition system is introduced which claims to recognize 10 different hand gestures in a car at day and night. This system uses a short-range radar together with a color camera and a depth camera. The sensors provide complementary information, which increases the accuracy when they are fused. The multi-sensor system is robust against varying lighting conditions.  

However, real-time applications for recognition tasks in certain computing environments such as mobile devices or in-vehicles have to meet some other requirements such as resource efficiency aside from having ultra high performing models \cite{howard2017mobilenets}. The computation resources of such applications are limited. Hence, there is a need for deep but more lightweight networks, which still provide good \mbox{accuracy \cite{kopuklu2019resource}}. 

Current research in this field is focusing mostly on creating architectures with fewer parameters and smaller size by satisfying computation limits of applications such as mobile devices or in-vehicles but still achieving high accuracy in recognition tasks. In SqueezeNet \cite{iandola2016squeezenet}, the squeeze layer compacts the features and saves a lot of parameters, still achieving AlexNet-level accuracy. Another strategy called depthwise separable convolution proposed in  \cite{chollet2017xception} generalizes the idea of separable convolutions. In MobileNet \cite{howard2017mobilenets} and ShuffleNet \cite{ DBLP:journals/corr/ZhangZLS17}, they used this strategy to reduce the computation cost while maintaining good performance. ShuffleNet \cite{DBLP:journals/corr/ZhangZLS17} however, utilizes group convolution and depthwise separable convolution in a novel method and adds the channel shuffle operation to gain the state-of-art results among light-weighted networks.

In this work, we have applied several resource efficient 3D CNN architectures for performance analysis and selected the resource efficient 3D-MobileNetV2 as the CNN model for our online recognition experiments. The performances have been analyzed for each RGB, depth and infrared modalities in our dataset separately in addition to their fusion analysis. Fusion analysis has been achieved with score level fusion using the scores computed for each modality.

\section{DriverMHG Dataset}
\label{sec:dataset}

\begin{figure}[t!]
	\centering
	\includegraphics[width = .45\textwidth, ]{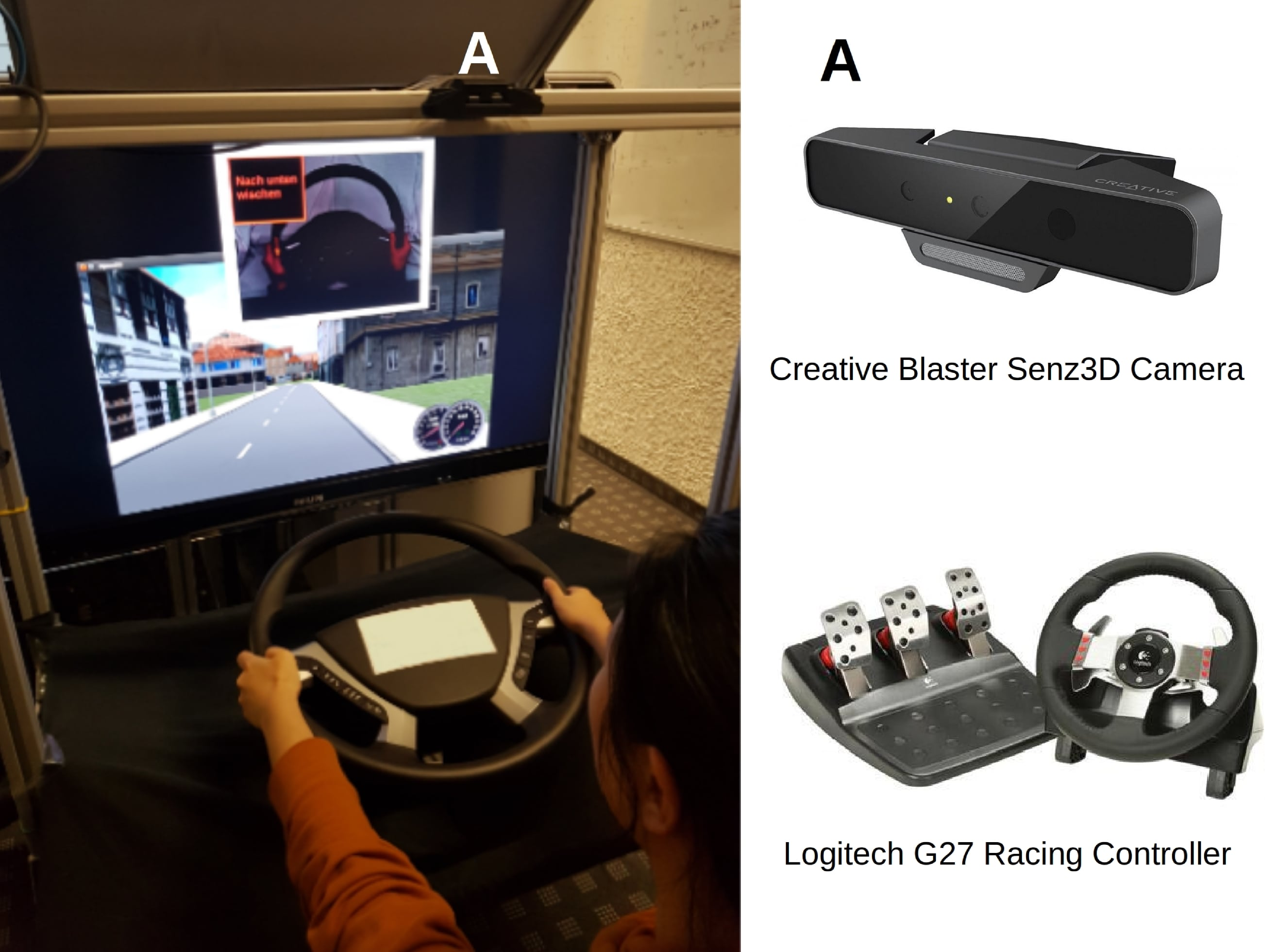}
	\caption{Driving Simulator - Setup. Left: The Complete Driving Simulator Setup showing a subject performing the driving task. Upper right: a picture of the mounted Creative Blaster Senz3D Camera. Lower Right: The Logitech G27 Racing Controller. }
	\label{fig:simulator}
\end{figure}

\begin{figure}[b!]
	\centering
	\includegraphics[width = 0.45\textwidth]{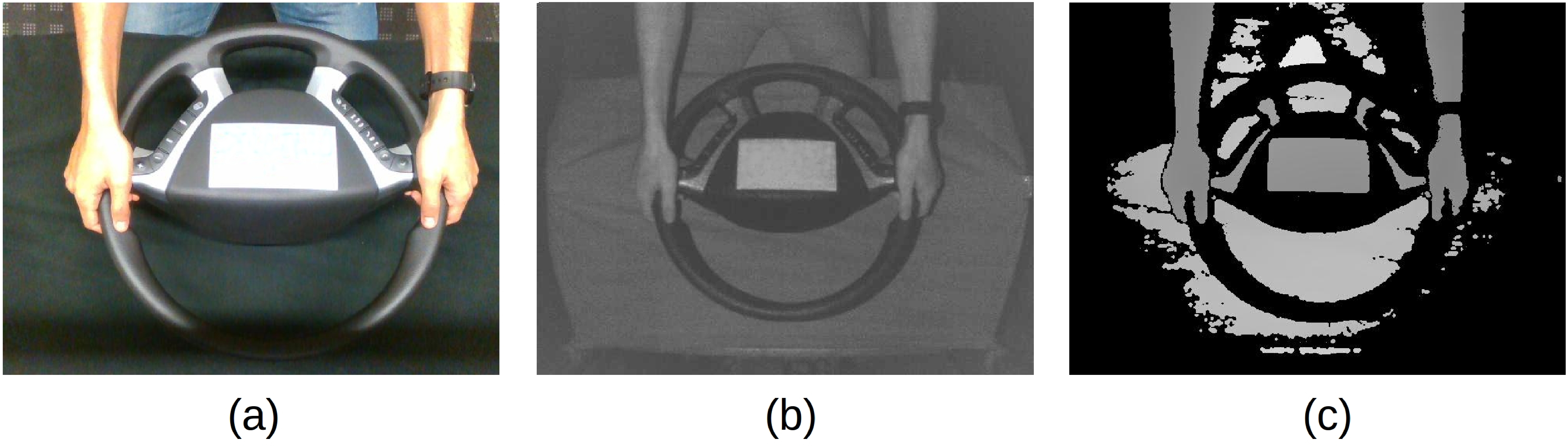}
	\caption{The dataset is collected for 3 different modalities: (a) RGB, (b) infrared, (c) depth.}
	\label{fig:dataset}
\end{figure}

There are a lot of vision-based datasets publicly available, but for the specific task of
classifying driver micro hand gestures on a steering wheel, there is none. For this purpose, we recorded the Driver Micro Hand Gesture (DriverMHG) dataset, which fulfills the following criteria:
\begin{itemize}
    \item Large enough to train a Deep Neural Network
    \item Contains the desired labeled gestures
    \item The distribution of labeled gestures is balanced
    \item Has '\textit{none}' and '\textit{other}' action classes to enable continuous classification
    \item Has the ability to allow benchmarking
\end{itemize}

In order to record this dataset, a driving simulator has been set up as shown in Fig. \ref{fig:simulator}. The dataset
was recorded with the help of 25 volunteers (13 males and 12 females) using this simulator, which consists of a monitor, a Creative Blaster Senz3D camera featuring Intel RealSense SR300 technology,  a Logitech G27 racing controller, whose wheel is replaced with a truck steering wheel and the OpenDS driving simulator software. The dataset is recorded in synchronized RGB, infrared and depth modalities with the resolution of \mbox{320 x 240} pixels and the frame rate of 30 fps. Example recordings for the three modalities are shown in Fig. \ref{fig:dataset}.

For each subject, there are in total 5 recordings each containing 42 gestures for 5 different gestures together with \textit{other} and \textit{none} gestures for each hand. Each recording of a subject was recorded under different lightning conditions: At room lights, at darkness, with external light source from left, with external light source from right and under intensive lightning from both sides. We randomly shuffled the order of the subjects and split the dataset by subject into training (72\%) and testing (28\%) sets. Recordings from subject 1 to 18 (including) belong to the training set, and recordings from subject 19 to 25 (including) belong to the test set.

\begin{figure}[t!]%
    \centering
    \subfigure[]{
    \includegraphics[width = 0.45\textwidth]{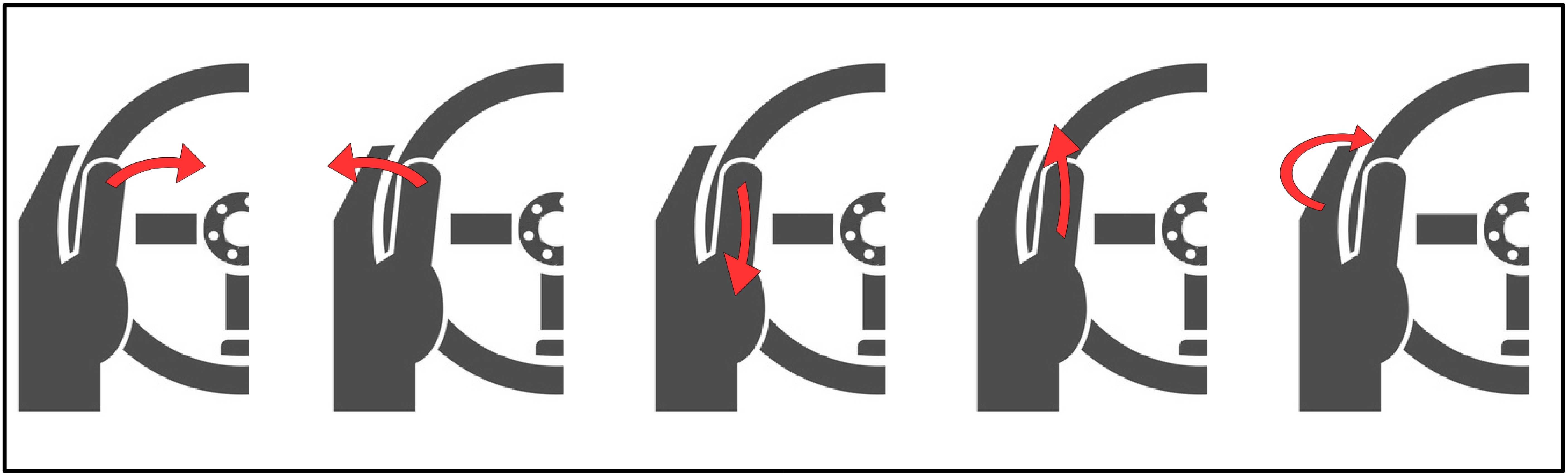}
    \label{fig:ges_left}}%
    \qquad
    \subfigure[]{
    \includegraphics[width = 0.45\textwidth]{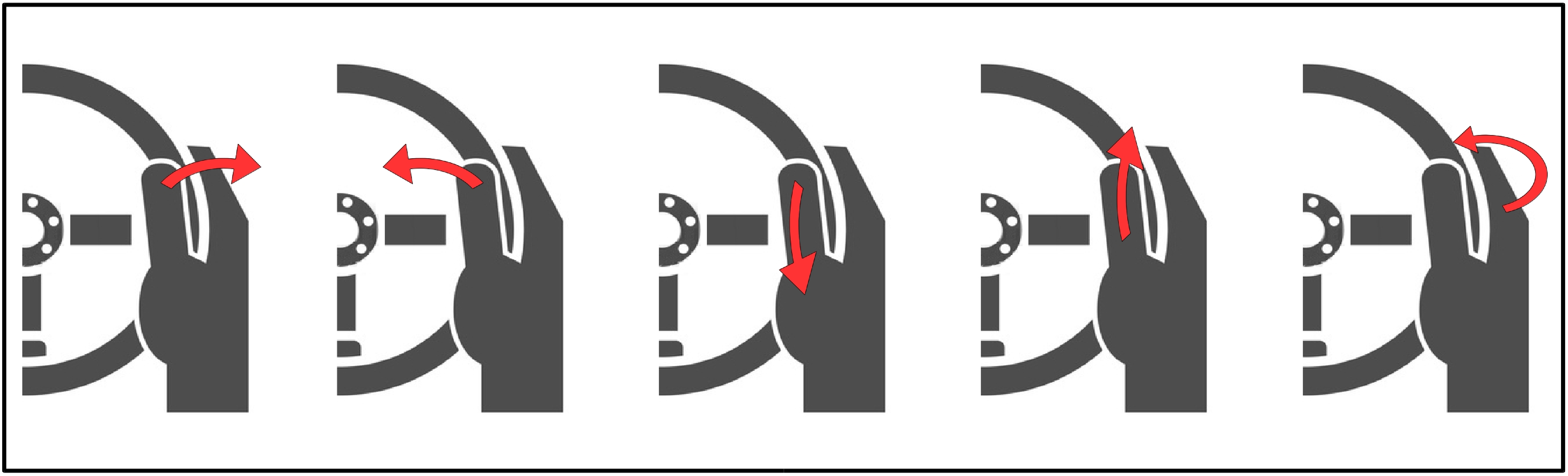}
    \label{fig:ges_right}}%
    \caption{Illustration of selected 5 micro gestures for (a) left and (b) right hands. From left to right: Swipe Right, Swipe Left, Flick Down, Flick Up and Tap. Besides these 5 gestures, \textit{none} and \textit{other} gesture classes are also collected for the DriverMHG dataset.}
    \label{fig:all_gestures}
\end{figure}

\begin{figure*}[t!]%
    \centering
    \subfigure[]{
    \includegraphics[width = 0.42\textwidth]{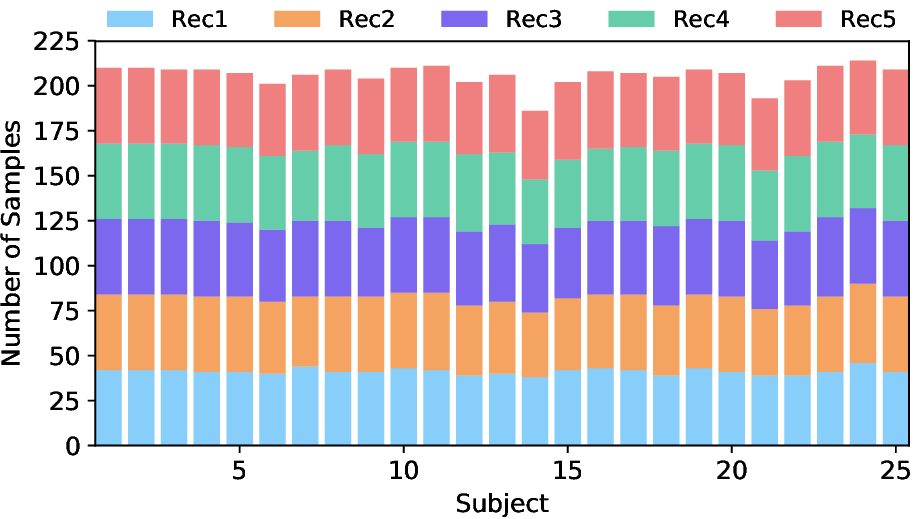}
    \label{fig:subject}}%
    \qquad
    \subfigure[]{
    \includegraphics[width = 0.42\textwidth]{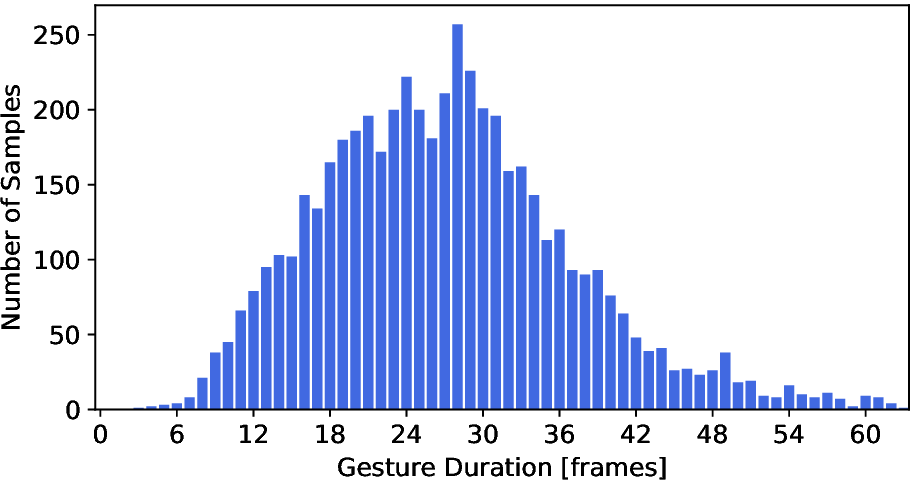}
    \label{fig:histogram}}%
    \qquad
    \subfigure[]{
    \includegraphics[width = 0.42\textwidth]{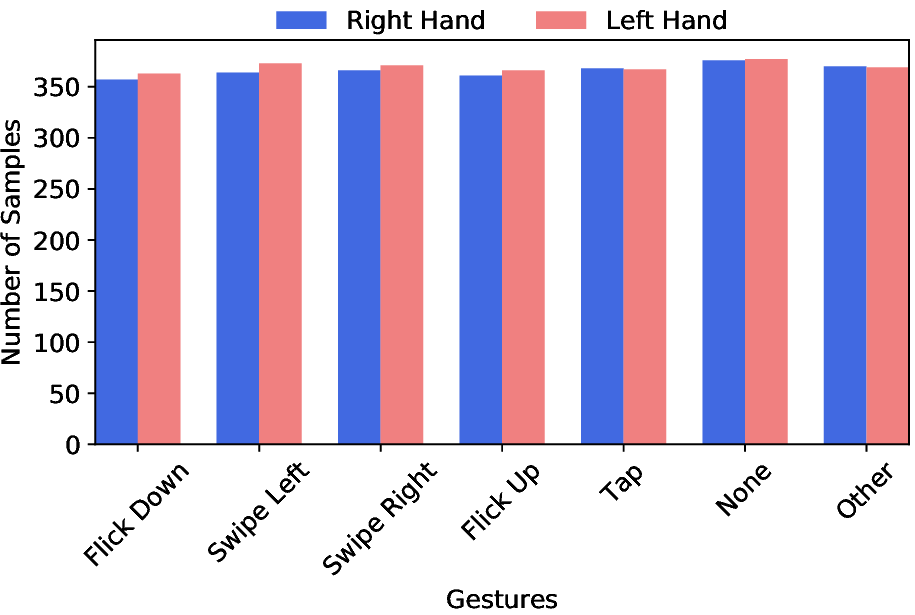}
    \label{fig:gesture_count}}%
    \qquad
    \subfigure[]{
    \includegraphics[width = 0.42\textwidth]{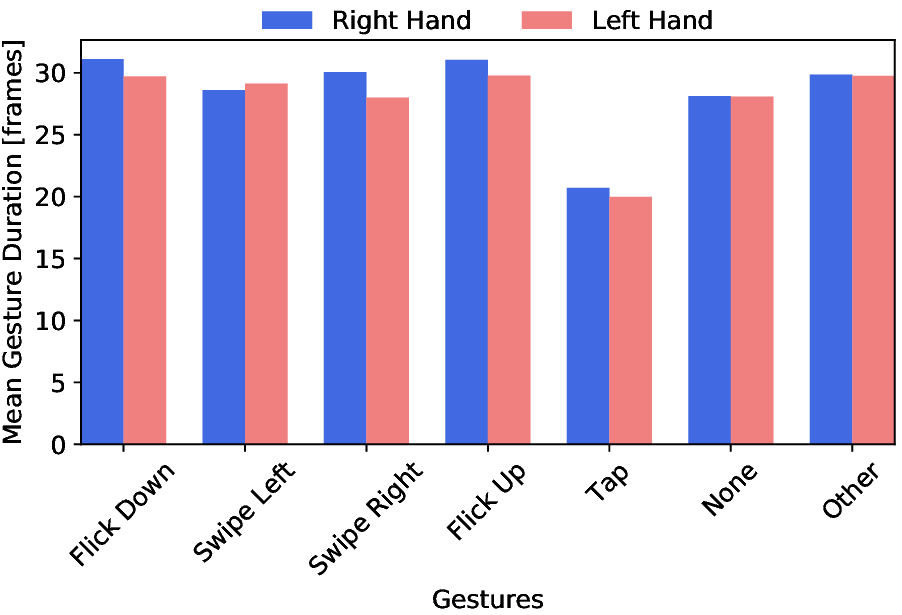}
    \label{fig:gesture_mean}}
    \caption{Statistics of the collected dataset: (a) Number of samples per subject, (b) Histogram of the duration of the gestures, (c) Number of samples per gesture, (d) Mean duration of gestures. }
    \label{fig:stanv}
\end{figure*}

The micro gestures that the subjects had to perform should be natural and should neither distract them nor require them to take their hands off the wheel while performing. They should also be quickly executable. Therefore, five micro gestures were selected, which are "\textit{swipe right}", "\textit{swipe left}", "\textit{flick down}", "\textit{flick up}" and "\textit{tap}". The former four gestures require the movement of only thumb, while "\textit{tap}" is performed by slightly patting the side of wheel with four fingers. Fig. \ref{fig:all_gestures} shows the illustration of the selected five micro gestures for the left and right hands. 

Additionally, we introduce the "\textit{other}" and "\textit{none}" gestures. For each record, three "\textit{none}" and "\textit{other}" gestures were specifically selected from the recorded data. With "\textit{other}" label, the network learns the drivers' other movements when a gesture is not performed. Whereas, with "\textit{none}" label, the network learns that the drivers' hands are steady (i.e. there is no movement or gestures). The inclusion of "\textit{none}" and "\textit{other}" action classes in the recorded dataset enables robust online recognition due to the availability of continuous analysis. Regarding the annotations, the gestures were annotated by hand with their starting and ending frame number.

Fig. \ref{fig:stanv} shows the statistics of the collected dataset. Fig. \ref{fig:stanv} (a) shows the number of samples from 5 recordings for each subject, which is around 210. Fig. \ref{fig:stanv} (b) shows the histogram of the gesture duration (frames). Fig. \ref{fig:stanv} (c) shows that the number of samples for each class are balanced for both right and left hands. In Fig. \ref{fig:stanv} (d), mean gesture duration for each action class are given. This figure shows that "\textit{tap}" action can be executed very fast compared to the other action classes. Mean gesture durations for "\textit{none}" and "\textit{other}" action classes are kept quite similar to the "\textit{flick down/up}" and "\textit{swipe left/right}" action classes.

\section{The Proposed Approach}
\label{sec:approach}

\begin{figure}[t!]
	\centering
	\includegraphics[width = 0.38\textwidth]{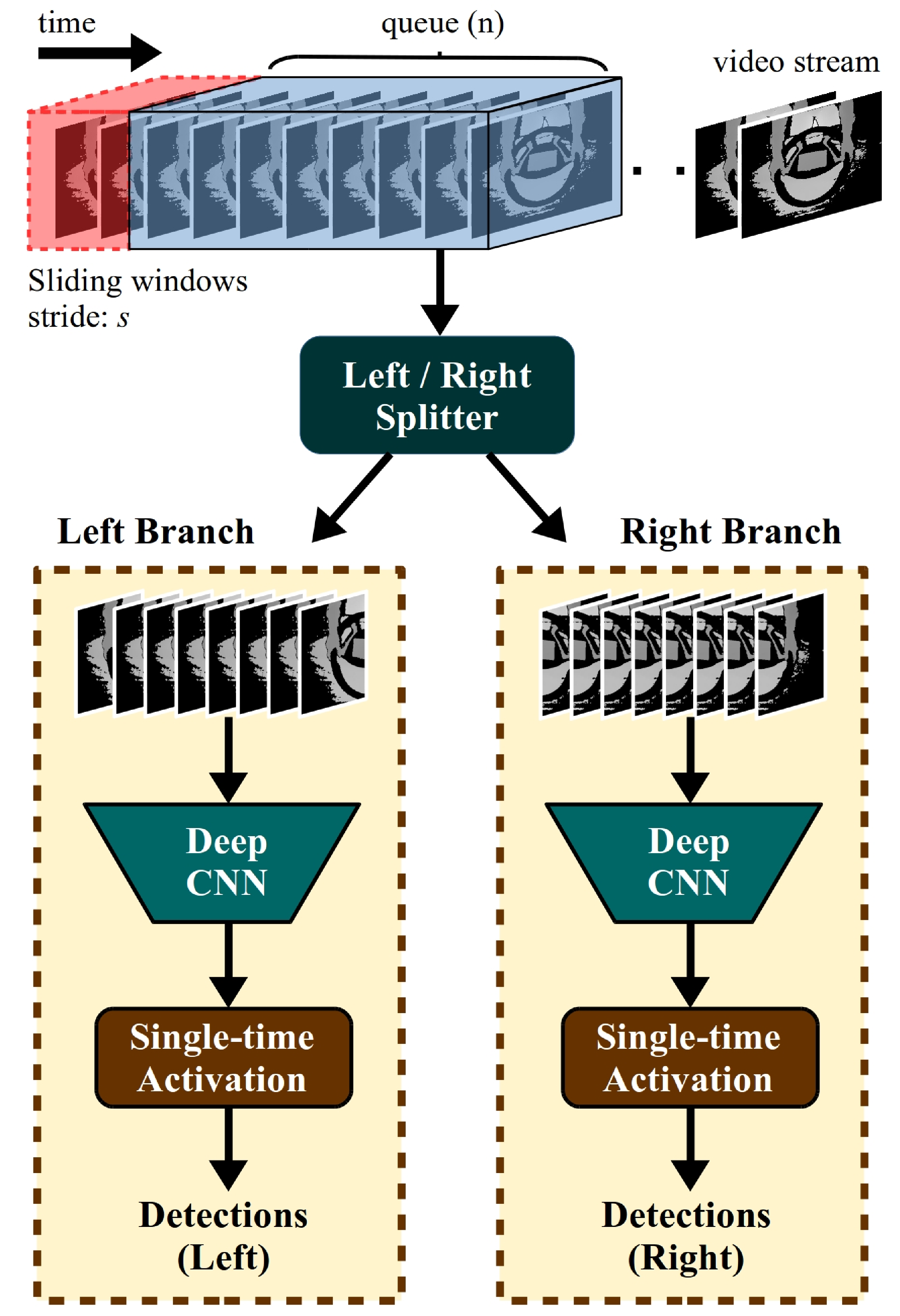}
	\caption{Online recognition architecture workflow. A sliding window holding \textit{n} frames is fed to a splitter. Splitter feeds the left and right halves of the clip frames to separate branches. Then, a resource efficient 3D-CNN model, which is trained separately for each branch, is applied to obtain class conditional scores of each gesture. Finally, \textit{single-time activation} block is applied to get detections.}
	\label{fig:arch}
\end{figure}

\subsection{Network Architecture}

The general network architecture is depicted in Fig. \ref{fig:arch}. For each modality, there is a queue holding the last \textit{n} frames of the video stream. Then, a video splitter is applied in order to separate the left and right halves of the video. The right and left halves contain the right and left hand related information, respectively. The reason of this splitting is to remove the unrelated video segment which behaves as a noise to the network. Afterwards, left and right video splits are fed to an offline trained CNN to get class conditional gesture scores for each hand. After this step, a single-time activation block is applied to get final detections.

\subsection{Offline Recognition}

Several resource efficient 3D-CNN models are used for offline trainings: 3D-MobileNet, 3D-MobileNetV2, 3D-ShuffleNet, 3D-ShuffleNetV2 and 3D-SqueezeNet. The details of these models can be found in \cite{kopuklu2019resource}. The left and right hands are trained separately, which gives the opportunity to recognize them independently. This way, simultaneous gestures from left and right hands can be recognized as tuples, which can be registered to an extra class as in \cite{kopuklu2019talking}. Training details for offline recognition are as follows:

\textbf{Learning:} Stochastic gradient descent (SGD) with standard categorical cross-entropy loss is applied for the trainings. For momentum and weight decay, $9 \times 10^{-1}$ and $1 \times 10^{-3}$ are used, respectively. For all trainings, pretrained models of the Jester dataset \cite{materzynska2019jester} are used. For modalities other than RGB, the initial convolutional layer's weights are altered accordingly. Specifically, weights across the RGB channels are averaged and replicated according to the number of input channels. Afterwards, we freeze all weights except for the last layer. The learning rate is initialized with $0.1$ and reduced twice with a factor of $10^{-1}$ after the validation loss converges. At the end, we fine tune all the parameters of the networks with learning rate of $0.001$ for 5 more epochs.

\textbf{Regularization:} Although the number of training samples for each gesture in the DriverMHG dataset is enough to train a deep CNN properly, over-fitting is still a serious problem since the number of gesture classes is small. Therefore, several regularization techniques are applied in order to achieve a better generalization. Weight decay of $\gamma = 1 \times 10^{-3}$ is applied to all parameters of the CNN models. A dropout layer is additionally inserted before the final linear or convolution layer with a dropout rate of 0.5. Moreover, data augmentation of multiscale random cropping is applied for all trainings.  

\textbf{Implementation:} The complete architecture is implemented and trained end-to-end in PyTorch.

\subsection{Online Recognition}

Online recognition is designed and evaluated for real driving scenarios. The test videos for online recognition from the DriverMHG dataset are unsegmented. Each test video has roughly 6500 frames and includes around 30 hand gestures. For online recognition, there are basically three requirements: (i) Detection of the starting/ending of the performed micro gestures, (ii) single-time recognition of the performed micro hand gestures and (iii) classification of the performed micro hand gestures. All these tasks above should be implemented in real time. Considering the aforementioned requirements, we propose \textit{Algorithm \ref{online recognition algorithm}} for online recognition, whose details are as follows.

\subsubsection{Detection of the starting/ending of the performed micro gestures}
It is essential to detect the starting and ending of a micro gesture for the created HCI system. In our proposed algorithm, there is no need for a separate detector, which saves a lot of computation and memory cost.

\begin{figure}[b!]
	\centering
	\includegraphics[width=0.45\textwidth]{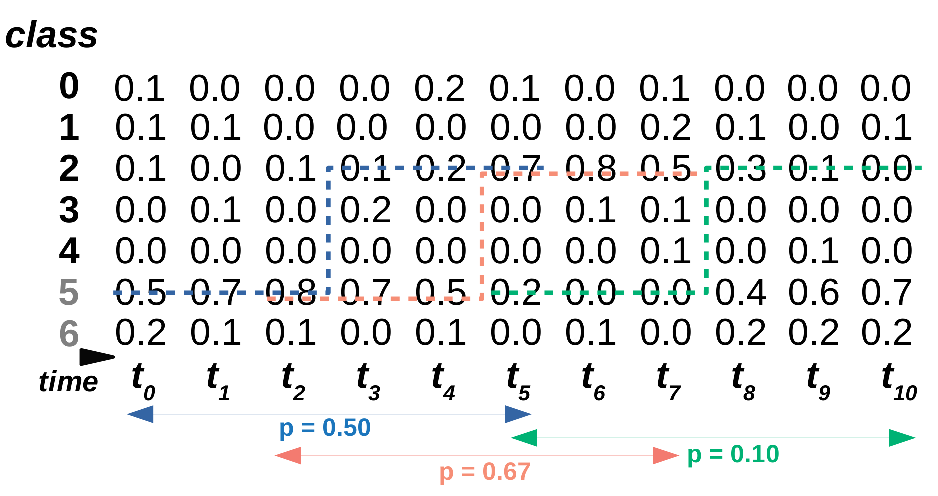}
	\caption{Illustration for the detection of the start and the end of a micro gesture. For the sake of simplicity, the length of sliding window is set to 6. Only three time instances and the transition from \textit{none} to gesture class 2 are depicted. Values 0.45, 0.62, 0.10 are the probabilities of detecting the transition pattern for the three time instances. Best in color.}
	\label{fig:transition detection}
\end{figure}

According to the recorded videos, the transition from \textit{none} or \textit{other} to any micro hand gesture represents the start of a micro gesture; on the contrary, the transition from any micro gesture to \textit{none} or \textit{other} represents the end of a micro gesture. To detect such a transition, we use a sliding window to calculate its probability as follows:

\begin{equation}
    p_{t,i} = \frac{ \overbrace{\sum_{n=t-l+1}^{t-\frac{l}{2}} P_{n,other/none}}^\text{none/other score} + \overbrace{\sum_{n=t-\frac{l}{2}+1}^{t} P_{n,i}}^\text{$i^{th}$ class score} }{l}
    \label{eq:trans_prob}
\end{equation}

\noindent where $i$, $l$ and $t$ denote the class type, window length and time step, respectively. $l$ is set to 64 in our experiments as it achieves the best performance. $P$ is the class conditional softmax score of the CNN output. The first half of Eq. (\ref{eq:trans_prob}) is the average score of \textit{other/none}, and the second half is the average score of $i^{th}$ class. Correspondingly, $p_{t,i}$ becomes the transition probability from \textit{other/none} to $i^{th}$ class at time instant $t$. A simplified example in Fig. \ref{fig:transition detection} helps to better understand the probability calculation. Class 5 and 6 are \textit{none} and \textit{other} classes, respectively.

\begin{algorithm}[!t]
  \caption{Online recognition algorithm}
  \label{online recognition algorithm}
  \begin{flushleft}
        \textbf{Input:} Frames in a "sliding window" strided by 1 over a test video\\
        \textbf{Output:} The sequence $\pi$ of the gestures inside the video
  \end{flushleft}
  \begin{algorithmic}[1]
    \For{each ``sliding window" with the length of $l$}
    \State{calculate $p_{t,i}$ for each class}
    \If{any $p_{t,i} > th_{s}$}
    \State{record classification outputs in $c$}
        \If{all $p_{t,i} < th_{e}$}
            \State{stop recording}
            \State{calculate the average classification result of $c$}
            \State{add the result to $\pi$}
    \EndIf
    \EndIf
    \EndFor
  \end{algorithmic}
\end{algorithm}

We set two hyperparameters $th_s$ and $th_e$ to indicate the start and the end of a micro gesture. For every time stamp, we calculate the transition from \textit{none} as well as from \textit{other} to each micro gesture (totally 10 possible paths). If any probability is larger than the threshold $th_s$, the micro gesture starts; if no probability is larger than $th_e$, the micro gesture ends. Those two thresholds can be different depending on different models.

\begin{table*}[t!]
    \centering
    \begin{tabular}{lcccccccccc}
        \specialrule{.15em}{.1em}{.1em}
        \multirow{4}{*}{\textbf{Model}} & \multirow{4}{*}{\textbf{MFLOPs}} & \multirow{4}{*}{\textbf{Params}} & \multicolumn{2}{c}{\multirow{2}{*}{\textbf{\underline{\phantom{nnnnn}Speed (cps)\phantom{nnnnn}}}}} & \multicolumn{6}{c}{\textbf{Accuracy (\%)}}   \\ \cmidrule(lr){6-11}
         &   &   & \multirow{2}{*}{\textbf{Titan XP}}   & \multirow{2}{*}{\textbf{Jetson TX2}}   & \textbf{Jester} & \multicolumn{3}{c}{\textbf{DriverMHG}} & \multicolumn{2}{c}{\textbf{DriverMHG (score fusion)}} \\ \cmidrule(lr){6-6} \cmidrule(lr){7-9} \cmidrule(lr){10-11}
         &   &   &   &   & \textbf{RGB} & \textbf{RGB} & \textbf{IR} & \textbf{D} & \textbf{RGB-IR} & \textbf{RGB-IR-D} \\
        \specialrule{.15em}{.1em}{.1em}
        3D-ShuffleNetV1 0.5x    & 76   & 0.27M  & 398  & 69  & 89.23  & 89.49  & 88.53   & 77.59  & 90.46  & 90.40  \\
        3D-ShuffleNetV2 0.25x   & 115  & 0.24M  & 442  & 82  & 86.91  & 87.83  & 87.90   & 73.21  & 89.08  & 88.39  \\
        3D-MobileNetV1 0.5x     & 97   & 0.88M  & 290  & 57  & 87.61  & 86.93  & 84.09   & 75.73  & 88.45  & 88.25  \\
        3D-MobileNetV2 0.2x     & 61   & 0.23M  & 357  & 42  & 86.43  & 88.18  & 85.68   & 76.49  & 89.22  & 88.46  \\
        \specialrule{.15em}{.1em}{.1em}
        3D-ShuffleNetV1 1.0x    & 198  & 0.97M  & 269  & 49  & 92.27  & 89.41  & 90.25   & 83.48  & 91.02  & 91.22  \\
        3D-ShuffleNetV2 1.0x    & 194  & 1.33M  & 243  & 44  & 91.96  & 88.93  & 90.12   & 77.32  & 90.46  & 90.60  \\
        3D-MobileNetV1 1.0x     & 240  & 3.33M  & 164  & 31  & 90.81  & 89.15  & 86.45   & 81.60  & 89.77  & 89.29  \\
        3D-MobileNetV2 0.45x    & 176  & 0.67M  & 203  & 19  & 90.21  & 89.56  & 89.15   & 83.61  & 90.39  & 90.74  \\
        \specialrule{.15em}{.1em}{.1em}
        3D-ShuffleNetV1 1.5x    & 345  & 2.01M  & 204  & 31  & 93.12  & 90.32  & 90.25   & 83.68  & 91.50  & 90.74  \\
        3D-ShuffleNetV2 1.5x    & 290  & 2.57M  & 186  & 34  & 93.16  & 89.49  & 90.46   & 80.71  & 91.08  & 90.60  \\
        3D-MobileNetV1 1.5x     & 427  & 7.34M  & 116  & 19  & 91.28  & 88.59  & 89.70   & 84.79  & 90.46  & 90.95  \\
        3D-MobileNetV2 0.7x     & 324  & 1.32M  & 130  & 13  & 93.34  & 90.95  & 89.08   & \textbf{86.93}  & 90.54  & 90.81  \\
        \specialrule{.15em}{.1em}{.1em}
        3D-ShuffleNetV1 2.0x    & 531  & 3.66M  & 161  & 24  & 93.54  & 90.95  & 90.74   & 83.89  & 91.78  & 91.78  \\
        3D-ShuffleNetV2 2.0x    & 436  & 5.47M  & 146  & 26  & 93.71  & \textbf{91.52}  & 91.23   & 83.67  & 92.40  & 91.92  \\
        3D-MobileNetV1 2.0x     & 660  & 12.93M & 88   & 15  & 92.56  & 89.15  & 89.08   & 84.59  & 90.81  & 90.87  \\
        3D-MobileNetV2 1.0x     & 559  & 2.39M  & 93   & 9   & \textbf{94.59}  & 91.36  & \textbf{91.56}   & 85.49  & \textbf{92.88}  & \textbf{92.47}  \\
        3D-SqueezeNet           & 922  & 1.85M  & 682  & 46  & 90.77  & 91.22  & 85.24   & 81.10  & 92.53  & 92.05  \\
        \specialrule{.15em}{.1em}{.1em}
    \end{tabular}
    \caption{Comparison of different 3D CNN architectures over offline classification accuracy, number of parameters, computation complexity (FLOPs) and speed on two different platforms. Models are trained separately for left and right hands and accuracies are calculated by dividing the number of correctly classified gestures to the total number of gestures. The calculations of parameters and FLOPs are done for the Jester dataset \cite{kopuklu2019resource} for 16 frames input with 112 $\times$ 112 spatial resolution. For the DriverMHG dataset, 32 frames input with 112 $\times$ 112 spatial resolution are used for trainings. For speed calculations (clips per second - cps), Titan Xp and Jetson TX2 platforms are used with batch size of 8.}
	\label{tab:comparison_offline}
\end{table*}

\subsubsection{Single-time recognition of the performed micro gestures}
The frames between the start and the end of a micro gesture are regarded as a valid clip. It is self-evident to pick the micro gesture with the highest score. This approach benefits from temporal ensembling, since it avoids the fluctuations of some false positive classifications. 

Combining the detection of a valid clip and single-time recognition, the algorithm of the whole online detection is described in \textit{Algorithm \ref{online recognition algorithm}}.
 
\subsubsection{Classification of the performed micro gestures}
To evaluate the performance of this online recognition algorithm, we use the Levenshtein distance to measure the difference between the output sequences and the ground truth sequences of input videos as in \cite{kopuklu2019real}. The Levenshtein distance represents the number of item-level changes, such as insertion, deletion, or substitutions. If the prediction is the same as the ground truth, the Levenshtein distance is then 0. The accuracy is 1 minus the fraction of the Levenshtein distance and the length of the ground truth sequence.

\section{Experiments and Results}
\label{sec:exp}

In this section, we evaluated the performances of our approach for the offline classification accuracy, including the impact of modality fusion on offline classification accuracy, and also for the online classification accuracy.

\subsection{Evaluation for Offline Classification Accuracy}

The recorded DriverMHG is evaluated for offline classifications accuracy using different types of resource efficient 3D-CNN architectures. However, although the number of training samples per class is sufficient to train a deep CNN, small number of classes leads to a relatively small dataset compared to other publicly available datasets, which lead to overfitting. Therefore we have initialized the weights of our models with Jester pretraining. Jester dataset is currently the largest available public dataset, which is a large collection of densely-labeled video clips that shows humans performing pre-defined hand gestures in front of
a laptop camera or webcam. It contains around 150,000 gesture videos. 

 \begin{figure}[b!]
	\centering
	\includegraphics[width = 0.48\textwidth]{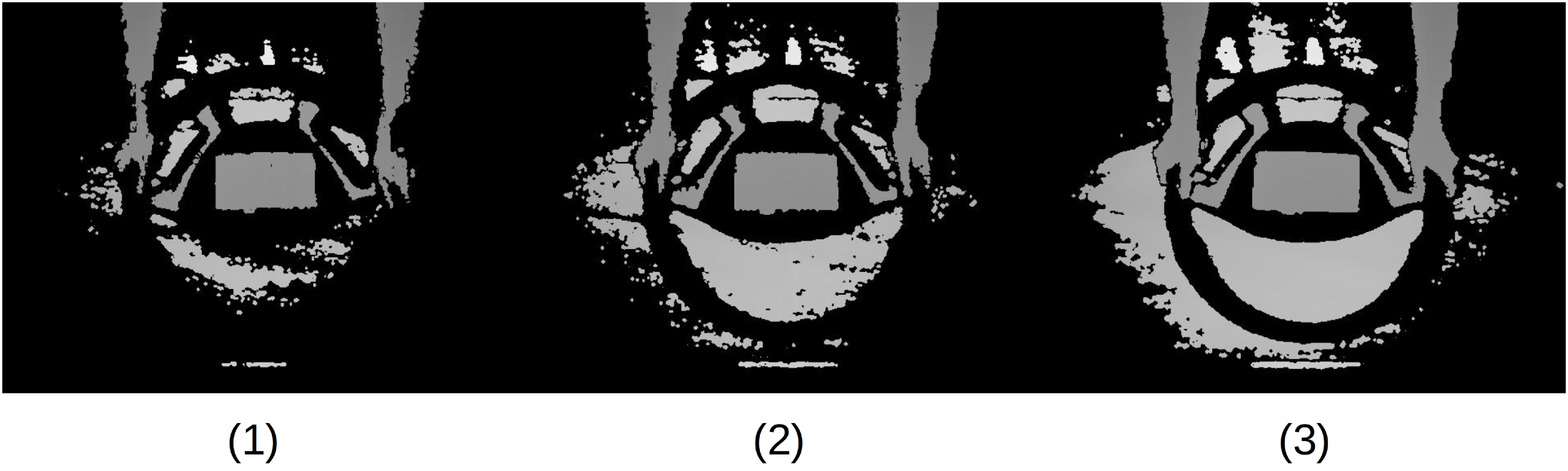}
	\caption{Depth images under different lightning scenarios: (1) With heavy solar radiation, (2) with medium solar radiation, (3) with light solar radiation.}
	\label{fig:depth_radiation}
\end{figure}

Table \ref{tab:comparison_offline} shows our offline classification accuracy results using both the Jester and the DriverMHG datasets. The models are trained separately for left and right hands and the average is reported in this table. The evaluations are done by applying five 3D resource efficient architectures, which are 3D-SqueezeNet, 3D-ShuffleNetV1, 3D-ShuffleNetV2, 3D-MobileNetV1 and 3D-MobileNetV2 for four complexity levels using the scaling factor \textit{width multiplier}. This scaling was not possible for the 3D-SqueezeNet hence the result is reported only for one complexity level for this architecture.

The applied approach provides very good offline classification accuracies on both datasets. The best classification accuracies are obtained with 3D-MobileNetV2 1.0x and 3D-ShuffleNetV2 2.0x for the Jester and DriverMHG datasets, respectively, on the RGB modality. 

In order to understand the performance of the proposed architecture on different modalities, we also evaluated offline classification accuracies with the infrared and depth modalities existing in our dataset. Table \ref{tab:comparison_offline} shows the results achieved for different modalities. Out of all modalities, infrared modality provides the best result (91.56\% achieved with 3D-MobileNetV2 1.0x) since it is invariant to illumination. Although, we expect similar results for the depth modality, it performs the worst due the inferior quality at intensive lightning conditions.

\begin{table}[t!]
     \centering
     \begin{tabular}{cc}
         \specialrule{.15em}{.1em}{.1em}
         \textbf{Temporal Dimension} & \textbf{Accuracy (\%)} \\ 
         \specialrule{.15em}{.1em}{.1em}
         16-frames (d:1)   &  85.96     \\
         16-frames (d:2)   &  89.84     \\
         16-frames (d:3)   &  86.99     \\
         32-frames (d:1)   &  \textbf{91.56}          \\
         \specialrule{.15em}{.1em}{.1em}
     \end{tabular}
     \caption{Comparative analysis of different temporal dimensions using model 3D-MobileNetV2 1.0x with IR modality.}
 	\label{tab:offline recogntion}
\end{table}

\begin{table}[t!]
     \centering
     \begin{tabular}{cc}
         \specialrule{.15em}{.1em}{.1em}
         \textbf{Modality} & \textbf{Levenshtein Accuracy (\%)} \\ 
         \specialrule{.15em}{.1em}{.1em}
         RGB  &  \textbf{74.00} \\
         IR   &  72.90 \\
         Depth    &  56.49 \\
         \specialrule{.15em}{.1em}{.1em}
     \end{tabular}
     \caption{Evaluation of the online detection using model 3D-MobileNetV2 1.0x.}
 	\label{tab:online recogntion}
\end{table}
 
Fig. \ref{fig:depth_radiation} shows an example for depth images under three lightning scenarios, which are under heavy solar radiation, under medium solar radiation and under light solar radiation. As it is clear from this figure, under intense lightning conditions, the quality of the depth modality drops significantly.

In Table \ref{tab:offline recogntion}, we also analyze the impact of different temporal dimensions for micro hand gesture recognition using the best performing model 3D-MobileNetV2 1.0x with infrared modality. Here, we tested the performances with 16 frames and 32 frames as input to the network and analyzed the impact of downsampling on recognition performances, as in \cite{kopuklu2018analysis}. The results show that when 16 frames are used for temporal analysis, downsampling by 2 performs the best. Downsampling 2 refers to uniform sampling of 16 frames out of 32 frames. For comparison purposes, we also selected 32 frames as temporal dimension and obtained the best recognition performance (91.56\%) without any downsampling. It must be noted that downsampling reduces number of floating point operations (FLOPs), which means faster runtime. The comparative analysis in Table \ref{tab:offline recogntion} shows that it is very critical to capture motion information in micro hand gesture recognition and this is why downsampling for temporal analysis does not help to improve results further. Fig. \ref{fig:stanv} (b) also shows that the frame duration of 32 is the correct duration for capturing micro hand gestures.

Table \ref{tab:comparison_offline} also reports the number of parameters, FLOPs and speed (clip per second - cps). As models get more complex, the number of parameters and FLOPs increase; as a corollary, the speed reduces. It must be noted that 3D-SqueezeNet is comparatively much faster although its FLOPs is the highest. This is due to the fact that it is the only architecture that does not make use of depthwise convolutions, and CUDNN library is specifically optimized for standard convolutions.  

\subsection{Impact of Modality Fusion on Offline Classification Accuracy}

In addition to the mono-modal analysis on RGB, infrared and depth modalities, we have analyzed the impact of fusion of multiple modalities on dynamic micro hand gesture recognition using the score level fusion strategy. 

According to the reported results in Table \ref{tab:comparison_offline}, the fusion of all three modalities enhances the offline classification accuracy with all the applied architectures for all complexity levels. However, another interesting result is that the fusion of RGB and infrared modalities performs better than the fusion of RGB, infrared and depth modalities. The reason is again poor quality of depth modality for intensive lightning scenarios, which degrades the fusion performance.

\begin{figure}[t!]
\begin{subfigure}
  \centering
  \includegraphics[width=\linewidth]{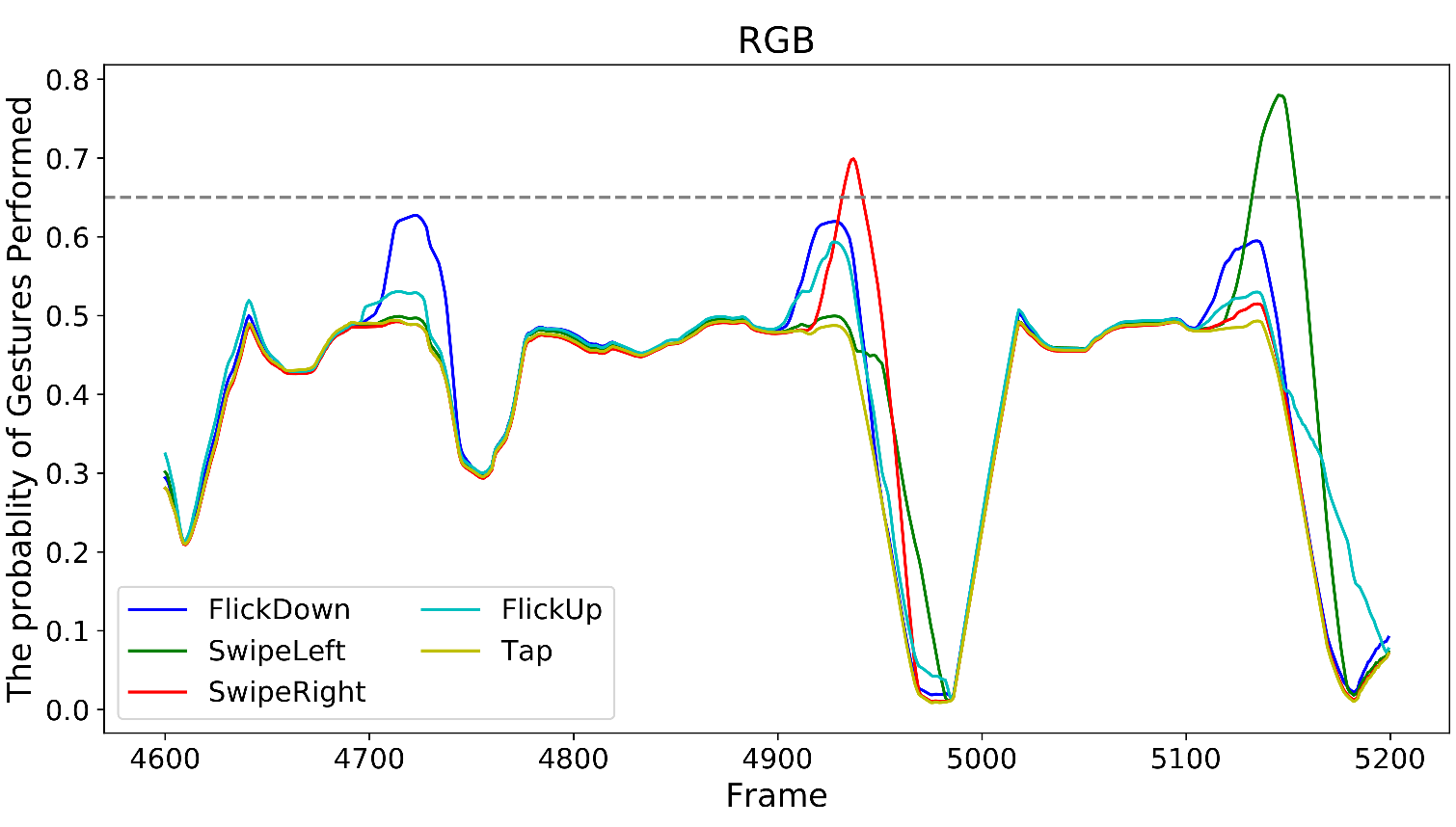}
  \label{fig:rgb}
\end{subfigure}
\begin{subfigure}
  \centering
  \includegraphics[width=\linewidth]{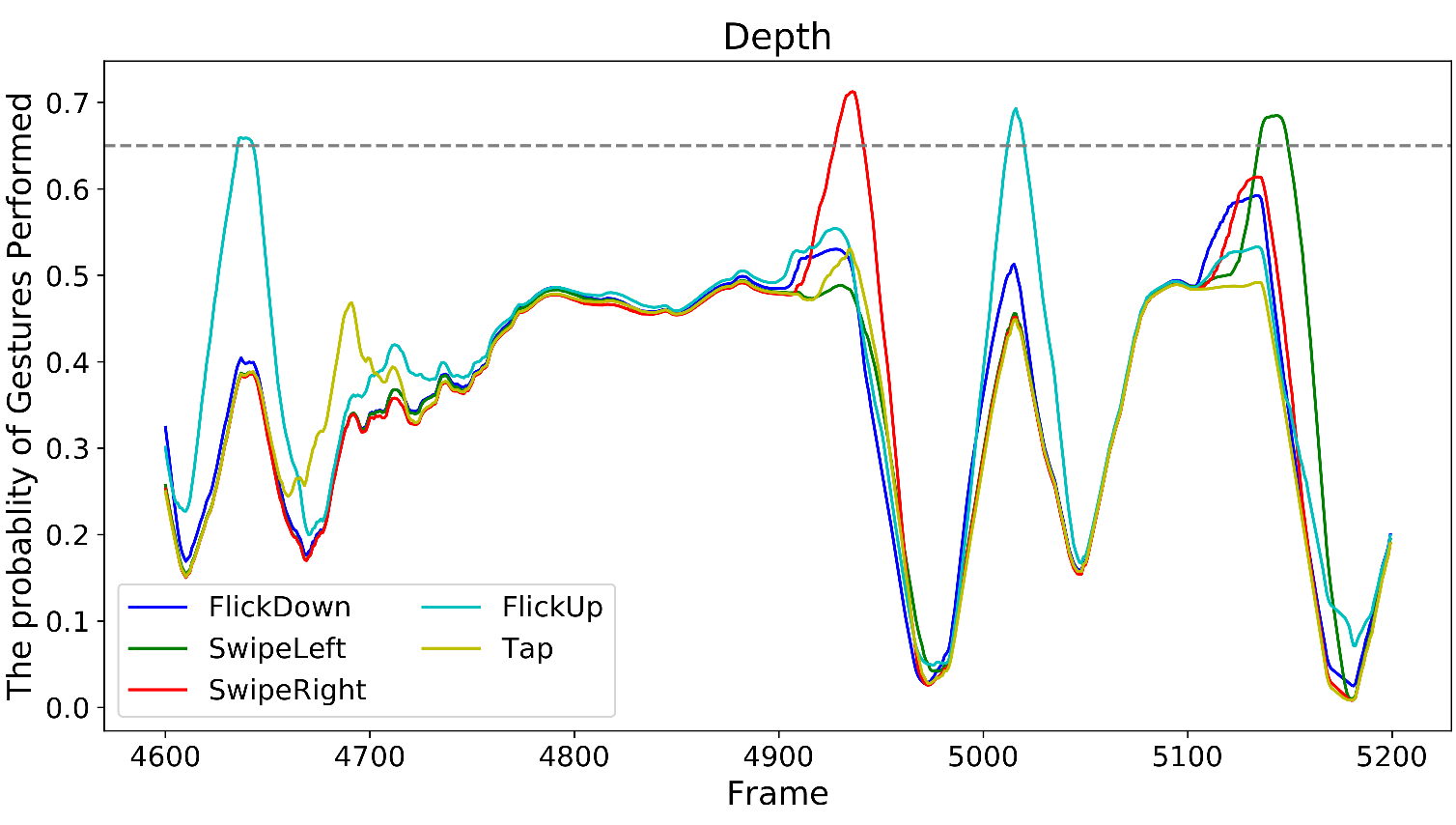}  
  \label{fig:d}
\end{subfigure}
\caption{Online recognition outputs in RGB (top) and depth (bottom) modalities for a test clip, which lasts 600 frames. The ground truth of this clip is \textit{Swipe Right} -- \textit{Swipe Left}. The gray dot line marks the threshold for the detection of gesture starting. The RGB outputs consist with the ground truth, while the depth outputs trigger two more false alarms. }
\label{fig:comparison}
\end{figure}

\subsection{Evaluation for Online Classification Accuracy}

Here, we use the 3D-MobileNetV2 as the CNN model in Fig. \ref{fig:arch} for the online evaluation, as its offline performance is in the lead according to the results reported in Table \ref{tab:comparison_offline}. 

Table \ref{tab:online recogntion} shows the online detection results for all RGB, infrared, and depth modalities. The best performance is achieved by modality RGB with 74\%. The results with infrared modality is slightly worse than the results with RGB modality. Depth modality provides a relatively poor performance, only 56.49\%. It is because that the depth images flicker too much, especially under intense lightning. Since the network is trained for recognizing micro gestures, the flickering (in a very small area) can result in false positives. 

Figure \ref{fig:comparison} shows the output of gesture detection in depth and RGB modalities for a short clip for comparison purposes. This figure explains the inferior performance of the depth images clearly.

\section{Conclusion}
\label{sec:conclusion}

This paper presents a new multi-modal dataset for dynamic recognition of driver micro hand gestures. This dataset was recorded from 25 subjects with a driving simulator and consists of three data modalities, which are RGB, infrared and depth. To the best of our knowledge, this is the first publicly available dataset that consists of micro hand gestures performed on a steering wheel.

Based on this novel dataset, we create an HCI system for online recognition of dynamic micro hand gestures. This kind of a system can have a crucial impact in automotive sector especially for safety related issues by preventing to perform some actions that can cause eyes off the road for few seconds. 

The performance analyses for offline classification have been achieved with several resource efficient architectures tested on three types of modalities existing in our dataset, which are RGB, infrared and depth modalities. Furthermore, in addition to mono-modal analysis, we have also analyzed the impact of fusion of different modalities on recognition performances. The results show that score level fusion of modalites enhances the recognition performances further. Out of all modalities, we were able to achieve the best results with infrared modality, which is robust to illumination changes. 

Since in automotive applications, the platform has limited resources and the reaction time is very crucial for the developed recognition technologies, in this paper, we propose a new HCI system considering both resource efficiency and fast reaction time in addition to achieving considerable recognition accuracy. However, we achieved 74.00\% online (Levenshtein) accuracy compared to 91.56\% offline accuracy, which shows that online recognition of micro gestures is challenging and open to improvements. We plan to improve the online recognition performance using different recognition strategies as a future work.


\section*{Acknowledgements}
We gratefully acknowledge the support of NVIDIA Corporation with the donation of the Titan Xp GPU used for this research.

{\small
\bibliographystyle{ieee}
\bibliography{egbib.bib}
}

\end{document}